\documentclass[a4paper,twoside]{article}

\usepackage{epsfig}
\usepackage{subcaption}
\usepackage{calc}
\usepackage{amssymb}
\usepackage{amstext}
\usepackage{amsmath}
\usepackage{amsthm}
\usepackage{multicol}
\usepackage{pslatex}
\usepackage{apalike}
\usepackage[bottom]{footmisc}
\usepackage{SCITEPRESS}     

\usepackage{cuted}
\usepackage{times}
\usepackage{url}
\usepackage{latexsym}
\usepackage{graphicx}
\usepackage{comment}
\usepackage{booktabs}
\usepackage{multirow}

\newtheorem{definition}{Definition}

\begin{document}

\title{Tag-Set-Sequence Learning for Generating Question-Answer Pairs}

\author{\authorname{Cheng Zhang and Jie Wang
}
\affiliation{Department of Computer Science, University of Massachusetts, Lowell, MA 01854, USA}
\email{cheng\_zhang@student.uml.edu, wang@cs.uml.edu}
}


%
\keywords{Tag-Set-Sequence Learning, Question-Answer Pairs, Natural Language Processing}


\abstract{
Transformer-based QG models can generate question-answer pairs (QAPs) with high qualities, 
but may also generate silly questions for certain texts. 
%
We present a new method called tag-set sequence learning to tackle this problem,
where a tag-set sequence is a sequence of tag sets to capture the syntactic and semantic information of the underlying sentence, and a tag set consists of one or more language feature tags, including, for example, 
semantic-role-labeling, 
part-of-speech, named-entity-recognition, and sentiment-indication tags. 
We construct a system called TSS-Learner to
learn tag-set sequences from given declarative sentences
and the corresponding interrogative sentences, and derive answers to the latter. 
We train a TSS-Learner model for the English language using a small training dataset and show that it can indeed generate adequate QAPs for certain texts that transformer-based models do poorly. 
Human evaluation on the QAPs generated by TSS-Learner over SAT practice reading tests is encouraging. 
}

\onecolumn \maketitle \normalsize \setcounter{footnote}{0} \vfill

\section{\uppercase{Introduction}}
\label{sec:introduction}

Multiple-choice questions (MCQs) are often used to
assess if students understand the main points of a given article.
An MCQ consists of an interrogative sentence, a correct answer (aka. answer key), and a number of
distractors.
A QAP is an interrogative sentence and its answer key. 
We study how to generate QAPs from declarative sentences.  

The coverage of the main points of an article may be obtained by selecting important declarative sentences using
a sentence ranking algorithm such as CNATAR (Contextual Network and Topic Analysis Rank) \cite{Zhang-Zhou-Wang2021} on the article. 
QAPs may be generated by applying a text-to-text transformer on a declarative sentence, a chosen answer key, and the surrounding chunk of sentences using, for example, TP3 (Transformer with Preprocessing and Postprocessing Pipelines) \cite{Zhang-Sun-Zhang-Wang2022}, which generates interrogative sentences for the answer keys with much higher success rates over previous methods.
TP3, however, may generate silly QAPs for certain chunks of texts. 

Training a deep learning model like TP3 for generating QAPs may be viewed as learning to speak in a language environment, akin to how kids learn to talk from their environment.
 On the other hand, learning to write well would require kids to receive formal educations. This analogy motivates us to explore machine-learning mechanisms that could mimic explicit rule learning. 

To this end we devise a  method using a tag-set sequence to represent the pattern of a sentence, where each tag set
consists of a few language-feature tags for the underlying word or phrase in the sentence. 
Given a declarative sentence and a corresponding interrogative sentence,
we would like to learn the tag-set sequences for each of these sentences and derive the tag-set sequence for the answer key, so that 
when we are given a new declarative sentence as input, we could generate a QAP by
first searching for a learned tag-set sequence
that matches the tag-set sequence of the input sentence,
 then use the learned tag-set sequence of the corresponding interrogative sentence
 to map the context of the input sentence to produce an interrogative sentence and
 derive a correct answer. 
 
We construct a general framework called TSS-Learner (Tag-Set-Sequence Learner) to
 carry out this learning task.
%
We train a TSS-Learner model for the English language over a small initial training dataset using  SRL (semantic-role-label), POS (part-of-speech), and
NER (named-entity-recognition)
tags,
where each data entry consists of a well-written
declarative sentence and a well-written interrogative sentence.
We
show that TSS-Learner can generate adequate QAPs for certain texts on which TP3 generates silly questions.  Moreover,
TSS-Learner can generate
 efficiently a reasonable number of 
adequate QAPs.
On QAPs generated from the official SAT practice reading tests, evaluations by human judges indicate that
97\% of the QAPs are both grammatically and semantically correct.

The initial training dataset is not big enough to contain tag-set sequences that match the tag-set sequences of a number of declarative sentences in the passages of the SAT reading tests. We manually add new  interrogative sentences
for some of these declarative sentences, and show that TSS-Learner is able to learn new rules and use them to generate additional adequate QAPs.

The rest of the paper is organized as follows: We summarize in Section \ref{sec:2} related work and present in Section \ref{sec:3} a general framework of tag-set-sequence learning. We then present in Section \ref{sec:4} an implementation of TSS-Learner for the English language and describe evaluation results in Section \ref{sec:5}. Section \ref{sec:6} concludes the paper.

\section{\uppercase{Related Work}} \label{sec:2}

Automatic question generation (QG), first studied by Wolfe \cite{wolfe1976automatic} as a means to aid independent study, has attracted much research
in two lines of methodologies with a certain success; they are transformative and generative methods.
 
\subsection{Transformative Methods}
Transformative methods transform key phrases from a single declarative sentence into interrogative sentences, including rule-based, semantics-based, and template-based methods.

Rule-based methods parse sentences using a syntactic parser to identify key phrases and transform a sentence to an interrogative sentence based on syntactic rules, including methods
to identify key phrases from input sentences and use syntactic rules for 
different types of questions  \cite{varga2010wlv}, 
generate QAPs using a syntactic parser, a POS tagger, and an NE analyzer \cite{ali2010automation},
transform a sentence into a set of interrogative sentences using a series of domain-independent rules \cite{danon2017syntactic}, and
generate questions using relative pronouns and adverbs from complex English sentences \cite{khullar2018automatic}. 

Semantics-based methods create interrogative sentences using
predicate-argument structures and semantic roles \cite{mannem2010question},
semantic pattern recognition  \cite{mazidi2014linguistic}, 
subtopics based on Latent Dirichlet Allocation \cite{chali2015towards}, or
semantic-role labeling 
\cite{flor2018semantic}.

Template-based methods are used for special-purpose applications with built-in templates, including
methods based on Natural Language Generation Markup Language (NLGML) \cite{cai2006nlgml},
on phrase structure parsing and enhanced XML
 \cite{rus2007experiments},
on self questioning \cite{mostow2009generating},
on enhanced self-questioning \cite{chen2009aist},
on pattern matching and templates similar to NLGML \cite{wyse2009generating},
on templates with placeholder variables \cite{lindberg2013automatic}, and
on semantics turned to templates \cite{lindberg2013generating}.

\subsection{Generative methods}
Recent advances of deep learning
have shed new light on
generative methods.
For example,
the attention mechanism  \cite{luong-etal-2015-effective} is 
used to determine what content in a sentence should be asked,
and the sequence-to-sequence  \cite{bahdanau2014neural,cho-etal-2014-learning} and the long short-term memory  \cite{Sak2014LongSM}  mechanisms are used to generate each word in an interrogative sentence  (see, e.g., \cite{du-etal-2017-learning,duan-etal-2017-question,Harrison_2018,sachan-xing-2018-self}).
These models, however, only deal with question generations
without generating correct answers. 
Moreover, training these models require a dataset comprising over 100K interrogative sentences.

To generate answers, researchers have explored ways to encode a passage (a sentence or multiple sentences) and an answer word (or a phrase) as input, and determine what interrogative sentences are to be generated for a given answer
\cite{10.1007/978-3-319-73618-1_56,zhao-etal-2018-paragraph,song-etal-2018-leveraging}.
Kim et al. \cite{Kim_2019} pointed out that these models could generate a number of
answer-revealing questions (namely, questions contain in them the corresponding answers). They then
devised a new method
by encoding answers separately, at the expense of having substantially more parameters. 
This method, however, suffers from low accuracy, and 
it is also unknown whether the generated interrogative sentences are grammatically correct. 

Recently, a new method is presented to perform a downstream task of transformers with preprocessing and postprocessing pipelines (TP3) for generating QAPs \cite{Zhang-Sun-Zhang-Wang2022}. They showed that
TP3 using pretrained T5 models \cite{raffel2020exploring} outperforms previous models. Human evaluations also confirm the high qualities of QAPs generated by this method. However, TP3 may generate silly questions for certain chunk of texts. This calls for further investigations for improving the qualities of generated QAPs.

\section{\uppercase{General Framework}} \label{sec:3}

Let $L$ be a natural language. Without loss of generality, we assume that $L$ has an oracle $O_L$ that can perform the following tasks:

\begin{enumerate}
\item  Identify simple sentences with no subordinate clauses and complex sentences with subordinate clauses.
\item Segment
a complex sentence into simple sentences for each clause.

\item Segment a sentence into a sequence of basic units, where
a basic unit could be a word, a phrase, or a subordinate clause.

\item Assign each basic unit in a sentence with one or more feature tags including, but not limited to,
POS, NER, SRL, and SEI (sentiment-indicator) tags.

\end{enumerate}

Existing NLP tools for
the English language, for example, provide a reasonable approximation to such an oracle. 

\subsection{Definitions}
\begin{definition}
Let $k \geq 2$ be a number of tags that $O_L$ can assign to a basic unit.
A $k$-tag set is a set of $k$ tags, 
denoted by $[t_1/t_2/\cdots/t_k]$ with a fixed order of tags: $t_1$ is an SRL tag, $t_2$ is a POS tag, $t_3$ is an NER tag, and $t_i$ ($i>3$) represent the other tags or a special word such as an interrogative pronoun.
\end{definition}


Two consecutive tag sets $A$ and $B$ with $A.1 = B.1$ (i.e., they have the same SRL tag)
and $A$ is left to $B$ in a sentence may be merged to a new tag set $C$ as follows:
(1) If $A = B$, then let $C \leftarrow A$.
(2) Otherwise, based on the underlying language $L$, either let $C.2 \leftarrow A.2$ (i.e., use the POS tag on the left) or let $C.2 \leftarrow B.2$.
For the rest of the tags in $C$, select a corresponding tag in $A$ or $B$ according to $L$. We have the following proposition:
%
For any sequence of tag sets, after merging, the new sequence of tag sets does not
have two consecutive tag sets with the same SRL tag.

\begin{definition} \label{def:2}
A tag-set sequence is a sequence  of interrogative pronouns (if any) and tag sets such that 
each SRL tag appears in at most one tag set.
\end{definition}

\noindent
\textbf{Remark.} To accommodate improper segmentation of phrasal verbs in applications, 
we may modify this definition of tag-set sequence by 
allowing a fixed number of consecutive tag sets to have the same SRL tag.

Since $O_L$ can segment
a complex sentence into simple sentences for each clause,
we treat such a sentence as 
a set of simple sentences. If a clause itself is a complex sentence, it can be further
segmented as a set of simple sentences. A declarative sentence consists of at least three different SRL tags corresponding to
subject, object, and predicate.

\subsection{Architecture}

\begin{figure}[h]
  \centering
  \includegraphics[width=\linewidth]{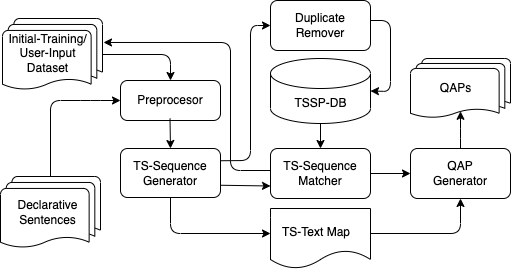}
  \caption{TSS-Learner architecture and data flow.}
  \label{fig:1}
\end{figure}

TSS-Learner learns tag-set sequences from a training dataset, where each data entry consists of
a simple declarative sentence and a interrogative sentence.
It consists of two phases: the learning phase and the generation phase. In the learning phase,
TSS-Learner learns tag-set sequence pairs from an initial training dataset to generate an initial 
TS-sequence pair database (TSSP-DB).
In the generation phase, it takes a declarative sentence as input and 
generates QAPs using TSSP-DB.
Figure \ref{fig:1} depicts the architecture and data flow of TSS-Learner,
which consists of five components: Preprocessor, TS-Sequence Generator, Duplicate Remover, 
TS-Sequence Matcher, and QAP Generator (see Section \ref{sec:4} for detailed explanations of these components in connection to an implementation of TSS-Learner for
the English language). 

\subsection{Learning phase}
The learn phase and the generation phase use the same Preprocessor and TS-Sequence Generator. Preprocessor is responsible for
creating tag sets for a given sentence and segmenting complex sentences into a set of simple sentences. TS-Sequence Generator is responsible for merging tag sets to form a tag-set sequence. Moreover, for an input sentence in the generation phase, it also maps each tag set after merging to the underlying text to be used later 
for generating QAPs. 


The Duplicate Remover component checks if a new 
pair of tag-set sequences generated by TS-Sequence Generator, one for a declarative sentence and the other
the corresponding interrogative sentence, is already in TSSP-DB. If yes, ignore it. Otherwise, deposit it in TSSP-DB.  




\subsection{Generation phase} \label{sec:generation}
Let $T$ be a tag-set sequence and
$T'$ the set of tag sets contained in $T$. Denote by $|T'|$ the size of $T'$.
Let $X_s$ be
the tag-set sequence for $s$ generated by TS-Sequence Generator, 
Recall that the text
for each tag set is stored in the TS-Text Map. 

Step 1. Find a tag-set sequence $X$ for a declarative sentence in TSSP-DB 
with the \textsl{best match} of $X_s$, meaning that the longest common substring of $X$ and $X_s$,
denoted by $Z= \text{LCS}(X,X_s)$, is the longest 
among all tag-set sequences in TSSP-DB.  A substring is a sub-sequence of consecutive tag sets.
If $Z$ contains tag sets for, respectively, a subject, a verb, and an object, then
we say that it is a \textsl{successful} match.
If furthermore, $Z = X = X_s$, then we say that it is  
a \textsl{perfect} match. If $Z$ is missing a subject tag set, a verb tag set, or
an object tag set, then it is an \textsl{unsuccessful} matching.
If a match is successful, go to Step 2.
If a match is unsuccessful or successful but not perfect, then notify the user that 
TSS-Learner needs to learn a new pattern and
ask for interrogative sentences for $s$ from the user. Go to Step 2.

Step 2.
Recall that $X$ is the best match with $X_s$ and there may be multiple pairs $(X,Y)$ in TSSP-DB because multiple interrogative sentences may be generated from the same declarative sentence.
Generate all possible interrogative sentences for $s$ as follows:  
For each pair of tag-set sequences $(X,Y) \in \text{TSSP-DB}$, 
generate a tag-set sequence $Y_s$ from $Y$ with
$$Y'_s = [Y'-(X' \cap Y'-X'_s)] \cup (X'_s - Z').$$
This means that $Y'_s$ is obtained from $Y'$ by removing tag sets that are in both
tag-set sequences in the matched pair $(X,Y)$,
but not in the input sentence,
and adding tag sets in the input sentence but not in $Z = \text{LCS}(X,X_s)$. We have 
$$X'_s - Z' = X'_s - X'.$$
Order tag sets in $Y'_s$  to form $Y_s$, which may require
localization according to the underlying language.
If a tag set in $Y'_s$ has the corresponding text stored in the TS-Text Map,
then replace it with the text. If not, then it would require localization to resolve it.
This generate an interrogative sentence $Q_{s}$ for $s$.

Step 3. For each interrogative sentence $Q_{s}$ generated in Step 2,
the tag sets in $A'_s = X' - Y'$ represent a correct answer.
Place tag sets in $A'_s$ in the same order as in $X'_s$ and replace each tag set with the corresponding text in $s$ to obtain an answer $A_s$ for $Q_{s}$.

\section{\uppercase{TSS-Learner for English}} 
\label{sec:4}

SRL, POS, and NER tags are used in this implementation. Existing
NLP tools for generating these tags are for words, not for phrases. 
Proper phrase segmentation can resolve this by merging.
In particular, it is critical to obtain segmentation for 
phrasal verbs for generating interrogative sentences.
A phrasal verb consists of a preposition or an adverb, or both. A straightforward method to segment phrasal verbs
is to use an extensive list of phrasal verbs.


\subsection{Essential NLP tools}
We use the following NLP tools  to generate tags: 
Semantic-Role Labeling \cite{shi2019simple} for SRL tags,
POS Tagging \cite{toutanova2003feature} for POS tags, and
Named-Entity Recognition \cite{peters2017semi}
for NER tags.

SRL tags are defined in PropBank\footnote{https://verbs.colorado.edu/~mpalmer/projects/ace/EPB-annotation-guidelines.pdf} \cite{bonial2012english,martha2005proposition}, 
which consist of three types: ArgN (arguments of predicates), ArgM (modifiers or adjuncts of the predicates) , and V (predicates).
ArgN consists of six tags: ARG0, ARG1, $\ldots,$ ARG5, and
ArgM consist of
multiple subtypes such as
LOC as location, EXT as extent, DIS as discourse connectives, 
ADV as general purpose, NEG as negation, MOD as modal verb, CAU as cause, TMP as time, PRP as purpose, MNR as manner, GOL as goal, and DIR as direction.

POS tags$\,$\footnote{https://www.cs.upc.edu/~nlp/SVMTool/PennTreebank. html} are defined in the Penn Treebank tagset \cite{toutanova2003feature,marcus1993building}. For example, NNP is for singular proper noun,
VBZ for third-person-singular-present-tense verb,
DT for determiner, and IN for preposition or subordinating conjunction.

NER tags include PER for persons, ORG for organization, LOC for locations, and numeric expressions for time, date, money, and percentage.

\begin{figure*}[t]
\textrm{Abraham [ARG1/NNP/PER/] Lincoln [ARG1/NNP/PER/] was [V/VBZ//] the [ARG2/DT//]
16th [ARG2/JJ//] president [ARG2/NN//] of [ARG2/IN//] the [ARG2/DT//] United [ARG2/NNP/LOC/]}
\caption{An example of a sentence and tag sets, where the tag set for each word is listed right after the word.}
\label{fig:example1}
\end{figure*}

\subsection{Preprocessor and TS-Sequence Generator}

On top of what is described in
Section \ref{sec:3}, Preprocessor 
%
%
%
first replaces contractions and slangs with
complete words or phrases to help improve tagging accuracy.
For example,  contractions \textsl{'m, 's, 're, 've, n't, e.g., i.e., a.k.a.} are replaced by,
respectively, \textsl{am, is, are, have, not, for example, that is, also known as}.
Slangs \textsl{gonna, wanna, gotta, gimme, lemme, ya}  are replaced by,
respectively, \textsl{going to, want to, got to, give me, let me, you}.
It then segments sentences and
tags words in sentences using SRL, POS Tagging, and NER
for the training dataset and later for input sentences for generating QAPs.
It also uses SRL to segment a complex sentence into a set of simple sentences and discards all simple sentences with
a subject or an object missing. 
Moreover, for each sentence, it removes
all the words with a CC (coordinating conjunction) as POS tag before its subject, including
\textsl{and, but, for, or, plus, so, therefore, because}.

TS-Sequence Generator merges the tag sets for words in each basic unit as follows:
(1) If the unit contains a V-tag set (i.e., a tag set with V as the SRL tag),
then use this V-tag set for the entire unit.
(2) If the unit contains no V-tag set, then it must contain a noun;
use the right most tag set that contains a noun POS tag.

It then merges the remaining tag sets if two consecutive tag sets are identical.
If they are not identical but have the same SRL tag,
then use this SRL tag in the merged tag set, and
the POS tag in the from the rightmost tag set. Moreover, the NER tag in the merged tag set is null
if both tag sets contain no NER tags; otherwise,
use the rightmost non-empty NER tag. 



\subsection{TS-Sequence Matcher}

This component takes a tag-set sequence $X_s$ of a sentence $s$ as input 
and executes Step 1 in the generation phase described in Section \ref{sec:generation}. The Suffix-Tree algorithm \cite{ukkonen1985algorithms} is used to compute a longest common substring of two
tag-set sequences, which runs in linear time. The POS tags  NN, NNP, NNS, and NNPS for various types of nouns
are treated equal. The POS tags VBP and VBZ
for third-person-singular-present verbs are treated equal.

Each tag set is in the form of $[t_1/t_2/t_3/t_4]$, where $t_1, t_2, t_3, t_4$ represent an SRL tag,
a POS tag, an NER tag, and an interrogative pronoun, and the latter two tags could possibly be null.
Fig. \ref{fig:example1} displays
the sentence ``Abraham Lincoln was the 16th president of the United States" and
the tag set for each word.


The resulting tag-set sequence for this sentence, after merging, is the following: \\
\textrm{[ARG1/NNP/PER/] [V/VBZ//] [ARG2/NNP/LOC/]}. 


\subsection{QAP Generator} 

QAP Generator executes Steps 2--3 in the generation phase described in
Section \ref{sec:3}. Recall that $Z = \text{LCS}(X,X_s)$ is the longest match
among all $(X,Y) \in \text{TSSP-DB}$. 
After $Y'_s$ is generated, we form $Y_s$ as follows:

\textsl{Case 1:} $Z=X_s$. Then $Y_s = Y$. 

\textsl{Case 2:} $Z$ is a proper substring of $X_s$. Then each tag set in $X'_s - Z$ appears either 
before $Z$ or after $Z$. Let $Y_b$ and $Y_a$ denote, respectively, the the tag set that appear before and after $Z$ in the same order as they appear in $X_s$.
Let $$Y_s =  [Y-(X'\cap Y'-X'_s)]Y_aY_b,$$ 
where $Y-(X'\cap Y'-X'_s)$
means to remove from $Y$ the tag sets in $X'\cap Y'-X'_s$.
 
For each tag set in $Y_s$, if a corresponding text can be found in the TS-Text Map,
then replace it with the text. A tag set that does not have a matched text in the
TS-Text Map is the extra helping verbs added to the interrogative sentence that generates $Y$.
There are five POS tags for verbs: VBG for gerund or present participle,
VBD for past tense, VBN for past participle, VBP for non-third-person singular present,
and VBZ for third-person singular present.
Present participle, past participle, and the negative forms of
past tense and present tense come with helping verbs. Thus,
only positive forms of the past tense VBD and the present tense VBP and VBZ do not come with helping verbs.

Suppose that there are two V-tag sets in $Y$, then the first V-tag set is for a help verb. If it does not have
a matching text in TS-Text Map, then it is an added helping word.
In this case, check the POS tag in the ARG0-tag set and determine if it is singular or plural. Then check the POS tag in the first V-tag set in $Y$ 
to determine the tense. Use the information of these POS tags to determine the correct form of the helping verb, and replace the second V-tag set with the matched verb in the TS-Text MAP in its original form. 

For example, suppose that the following declarative sentence ``John traveled to Boston last week" and its
interrogative sentence about location ``Where did John travel to last week"
are in the training dataset. They have the following tag sets before merging: \\
\\
``John [ARG0/NNP/PER/] traveled [V/VBD//] to [ARG1/IN//] Boston [ARG1/NNP/LOC/] last [TMP/NN//] week [TMP/NN//]", and
``Where [///where] did [V/VBD//] John [ARG0/NNP/PER/] travel [V/VB//] to [ARG1/IN//] 
last [TMP/NN//] week [TMP/NN//]".

With proper phrase segmentation we know that ``travel to" is a phrasal verb. Thus, after merging, we have
``John [ARG0/NNP/PER/] traveled  to [V/VBD//] Boston [ARG1/NNP/LOC/] last week [TMP/NN//]", and
``Where [///Where] did [V/VBD//] John [ARG0/NNP/PER/] travel to [V/VB//] last week [TMP/NN//]".

After merging, the pair of tag-set sequences $(X,Y)$ is deposited in TSSP-DB, where
$X =$
\textrm{``[ARG0/NNP/PER/] [V/VBD//] [ARG1/NNP/LOC/] [TMP/NN//]"}, and $Y = $
\textrm{``[///where] [V/VBD//] [ARG0/NNP/PER/] [V/VB//] [TMP/NN//]"}.

Suppose that we are given a sentence $s=$ ``Mary flew to London last month." Its tag-set sequence $X_s$ is exactly the same as $X$, with
[ARG0/NNP/PER/] for ``Mary",
[V/VBD//] for ``flew to", [ARG1/NNP/LOC/] for ``London", and [TMP/NN//] for ``last month", which
are stored in the TS-Text Map.
Thus, $Y_s = Y$. 
We can see that 
the tag set of [V/VB//] 
in $Y$ is not in the TS-Text Map and so cannot be matched.
To resolve this, check the POS tag in the ARG0-tag set, which is NNP, indicating a singular noun.
The POS tag in the first V-tag set is VBD, indicating past tense. Thus, the correct form of the helping verb is ``did". The text for [V/VBD//] is ``flew to" in the TS-Text Map. The original form of the verb is ``fly".
Thus, the second V-tag set is replaced by ``fly". 
This generates the following interrogative sentence: ``Where did Mary fly to last month?"
The tag set for the answer  is $X'-Y'$, which is [ARG1/NNP/LOC/], corresponding to ``London" in the TS-Text Map.

\begin{table*}[t]
\caption{Examples that TSS-Learner does well but TP3 does not, where
Q-by-TP3 and Q-by-TSSL stand for, respectively, ``Question generated by TP3" and ``Question generated by TSS-Learner". Note that in Example 3, in the input context, Eudora Welty took a couple of friends to a restaurant 
and an unknown woman came and joined them.}
\label{table:TP3}
\begin{tabular}{ll} 
\text{Example 1} & \\\hline
\text{Input text}: &``However, on September 12, 1933, physicist Leo Szilard invented the neutron-
chain reaction." \\
\text{Q-by-TP3}: &``When did the 11th of September happen?" \\
\text{Q-by-TSSL}: &``When did physicist Leo Szilard invent the neutron-induced nuclear chain reaction?" \\
\text{Answer}: &``on September 12, 1933"\\\hline
\\
\text{Example 2} & \\\hline
\text{Input text}: &``My mother bought the beautiful basket and put it safely in some hiding place I couldn't find." \\
\text{Q-by-TP3}: &``What did my mother say I needed?" \\
\text{Q-by-TSSL}: &``What did my mother buy?" \\
\text{Answer}: &``the beautiful basket"\\\hline
\\
\text{Example 3} & \\\hline
\text{Input text}: &``Without a second thought, the woman joined the Welty party." \\
\text{Q-by-TP3}: &``What party did Eudora Welty join without thinking about it?" \\
\text{Q-by-TSSL}: &``What did the woman join?" \\
\text{Answer}: &``the Welty party"\\\hline
\end{tabular}
\end{table*}

\section{\uppercase{Evaluations}} \label{sec:5}

\subsection{Training dataset}

We construct an initial training dataset by composing 112 pairs of declarative sentences
and the corresponding interrogative sentences to cover the common
tense, participles, voice, modal verbs, and some common phrasal verbs such as
``be going to" and ``be about to" for the following six interrogative pronouns: 
\textsl{Where, Who, What, When, Why, How many}. A total of 112 tag-set-sequence pairs are learned and deposited in the initial 
TSSP-DB.

Note that SQuAD \cite{Rajpurkar_2016} is a  dataset commonly
used for training and evaluating generative methods for QG.
However, a certain number of QAPs in SQuAD are formed improperly or lack correct answers. 
There are also about 20\% of questions in the dataset that require paragraph-level information.
Thus, SQuAD is unsuitable to train a TSS-Learner model.

\subsection{Comparisons with TP3}

Recall that
TP3 \cite{Zhang-Sun-Zhang-Wang2022} takes a declarative sentence with a specified answer key contained in it and the surrounding declarative sentences as input, TP3 generates QAPs with better qualities then previous methods, but it also generates silly QAPs for certain chunks of text. It would be interesting to know whether
TSS-Learner may be used to complement TP3. 
For this purpose we use the same dataset used to evaluate TP3, 
and we show that the trained TSS-Learner model with the small initial training dataset can indeed generate adequate QAPs for certain texts that TP3 does poorly. 
Table \ref{table:TP3} depicts some of these results.

\begin{table*}
\centering
\caption{Human evaluation results on the SAT practice reading tests.}
\label{table:human}
\begin{tabular}{l|c|c|c|c|c|c|c}
\hline
& \text{Where} & \text{Who} & \text{What} & \text{When} & \text{Why} & \text{How} & \text{Total} \\
\hline
TSSP-DB entry pairs & 18 & 45 & 23 & 22 & 6 & 8 & 122 \\
\hline
QAPs generated & 26 & 216 & 466 & 51 & 15 & 22 & 796 \\
\hline
All correct & 21 & 208 & 458 & 51 & 15 & 20 & 773 \\
\hline
Syntactically acceptable & 4 & 4 & 3 & 0 & 0 & 2 & 13 \\
\hline
Semantically acceptable & 1 & 2 & 5 & 0 & 0 & 0 & 8	 \\
\hline
Unacceptable & 0 & 2 & 0 & 0 & 0 & 0 & 2 \\
\hline
\end{tabular}
\end{table*}

\subsection{Human evaluation}

Computed metrics such as BLUE \cite{10.3115/1073083.1073135}, ROUGE \cite{lin-2004-rouge}, and Meteor \cite{10.1007/s10590-009-9059-4} are commonly used to evaluate summarization and machine translation against benchmark datasets, which  have also been used to evaluate the qualities of machine generated QAPs. 
These metrics, however, do not evaluate grammatical correctness, and so human judgments are needed

We asked three human judges to evaluate the qualities of the QAPs on a shared document based on the following criteria:
For questions: Check both grammar and semantics: (1) correct; (2) acceptable (i.e., a minor fix would make it correct); (3) unacceptable.
 For answers: (1) correct; (2) acceptable; (3) unacceptable.
 If judges have discrepancies on an item, they resolved it through discussions. In so doing, 
 they jointly produced one final evaluation result for each QAP. 

We use the official SAT practice reading tests$\,$\footnote{https://collegereadiness.collegeboard.org/sat/practice/full-length-practice-tests} for evaluating TSS-Learner.
These tests provide a large number of different patterns of declarative sentences in the underlying passages.
 There are a total of eight SAT practice reading tests, where each test consists of five articles and
 each article consists of around 25 sentences. There are 40 articles and 1,136 sentences. 
After removing interrogative sentences and other non-declarative sentences,
there are 1,025 declarative sentences.
Using the initial training dataset of 112 pairs of declarative and interrogative sentences, TSS-Learner generates a total of 796 QAPs.
Table \ref{table:human} depicts the evaluation summaries with detailed breakdowns in 
each interrogative pronoun, where ``all correct" means that the question in the QAP is both grammatically and semantically correct, conforms to what native speakers would say, and the answer is correct;
``syntactically acceptable" means that either the question or the answer has a minor grammar error, and a small effort such as changing, removing, or adding one word will fix the problem;
``semantically acceptable" means that either the question or the answer is problematic in semantics, but a minor effort will fix problem; and
``unacceptable" means that either the question or the answer is unacceptable grammatically or semantically.

The percentage of generated questions that are both syntactically and semantically correct
is 97\%. 
We noticed that there is a strong correlation between
the correctness of the questions and their answers. In particular, 
when a generated question is all correct, its answer is also all correct.
When a question is acceptable, its answer may be all correct or acceptable.
Only when a question is unacceptable, its answer is also unacceptable.
 

The 13 syntactically acceptable questions are mostly due to some minor issues
in segmenting a complex sentence into simple sentences, where a better
handling of sentence segmentation is expected to correct these issues. Two questions whose
interrogative pronoun should be ``how much" are mistakenly using ``how many".
Further refinement of POS tagging that distinguish uncountable nouns from countable nouns would solve this problem. The eight semantically acceptable questions are all due to 
NER tags that cannot distinguish between persons, location, and things. Further refinement of NER tagging will solve this problem. The two unacceptable questions are due to
serious errors induced when segmenting complex sentences. This suggests that we may need to 
find a better way to deal with complex sentences. Using a recursive list to represent complex sentences
might be useful in this direction.

There are 589 sentences for which no matched tag-set sequences are found from 
the initial training dataset. By learning new tag-set sequences from user inputs,
535 of these sentences found perfect matching, which generate
QAPs that are both syntactically and semantically correct. For the remaining 84 sentences,
it is hard to segment them into a set of simple sentences and so no
appropriate tag-set sequences were learned. This suggests that we should
look into better sentence segmentation methods or explore tag-set trees as recursive lists of tag-set sequences to represent complex sentences
as a whole for future studies.

\subsection{Efficiency}
Finally, we evaluate the running time for our trained TSS-Learner to generate QAPs over 100 sentences on a desktop computer with an Intel Core I5 2.6 Ghz CPU and 16 GB RAM. The average running time is 0.55 seconds for each input sentence, which is deemed satisfactory for online applications. For a given article, assuming that it would take the reader several minutes to read. By then all the QAPs for MCQs would have been generated.

\section{\uppercase{Conclusions}} \label{sec:6}

Tag-set sequence learning is a novel attempt for generating adequate QAPs. It can generate adequate QAPs that text-to-text transformers fail to generate. Numerical analysis shows that this approach is promising. It
achieves satisfactory results for the English language using existing NLP tools on
SRL, POS, and NER tagging. Further improvement of NER tagging may be able to eliminate a small number of semantic errors we encountered. 

Tag-set sequence learning is a text-to-rule learning mechanism and a text-to-text model via explicit rules. These explicit rules are automatically learned with moderate human involvement -- users are expected to write and provide to the system an interrogative sentence when a declarative sentence has a unseen tag-set sequence. This procedure continues to enrich the collection of tag-set sequence pairs in TSSP-DB,
When almost all possible patterns of declarative sentences and the corresponding interrogative sentences are learned (there are only finitely many of them to be learned), TSS-Learner is expected to perform well on generating adequate QAPs from declarative sentences that can be segmented appropriately into simple sentences.

However, not all complex sentences can be segmented using
existing tools. In particular, about 7.4\% of the declarative sentences in the 
official SAT practice reading tests are in this category. This calls for, as mentioned
near the end of Section \ref{sec:5}, a better NLP method to dissect complex sentences. 

Applying TSS-Learner to a logographic language would require robust and accurate segmentation at all levels of words, phrases, and sentences, semantic labeling, POS tagging, and named-entity recognition for the underlying languages.
It would also require appropriate  localization for
merging tag sets. 
It is interesting to explore how well TSS-Learner performs on a language other than English.
It is also interesting to investigate whether other NLP tools that better represent the structures of sentences,
such as dependency trees and constituency trees, can be combined with tag-set sequences to generate QAPs with a higher quality.


\bibliographystyle{apalike}
{\small
\bibliography{QG-reference}}

\end{document}